\documentstyle[ijcai95,fleqn,leqno]{article}

\newcounter{example}

\newenvironment{ex}{\refstepcounter{example}
\begin{description}\item[(\theexample)]}{\end{description}}

\newcommand{\bex}{\begin{ex}}
\newcommand{\eex}{\end{ex}}

\title{\bf Integrating Gricean and Attentional Constraints}

\author{\bf Rebecca J. Passonneau\thanks{The work reported here
was not supported by Bellcore.} \\
Bellcore \\
445 South Street \\
Morristown, NJ 07960  \\
beck@bellcore.com}

\begin{document}

\maketitle
\begin{abstract}
This paper concerns how to generate and understand
discourse anaphoric noun phrases.
I present the results of an
analysis of all discourse anaphoric noun phrases (N=1,233) in a corpus of ten
narrative monologues, where the choice between a definite
pronoun or phrasal NP conforms largely to
Gricean constraints on informativeness.  I discuss Dale~\&
Reiter's~\cite{dale&reiter94} recent model and show how it can be
augmented for understanding as well as generating the range of data
presented here.  I argue that integrating
centering~\cite{gjw83}~\cite{kameyama85} with this
model can be applied uniformly to discourse anaphoric pronouns and phrasal
NPs.  I conclude with a hypothesis for
addressing the interaction between local and
global discourse processing.
\end{abstract}

\pagestyle{empty}

\section{Introduction}
\label{intro}

This paper concerns how to generate and understand discourse anaphoric
noun phrases, or noun phrases (NPs) that evoke a discourse entity
already in the discourse model (Webber~\shortcite{webber78}).
Dale~\shortcite{dale89}~\shortcite{dale92} implements Gricean
constraints on informativeness for generating discourse anaphoric NPs.
However, his model follows the tradition of assuming that distinct
constraints govern pronouns versus phrasal NPs
(cf.~\cite{reichman85}~\cite{gs86}).
Centering~\shortcite{gjw83}~\shortcite{kameyama85}, a model of local
attentional state~\shortcite{sidner79}, has been applied primarily to
definite pronouns.  I argue that Gricean constraints should be applied
equally to discourse anaphoric pronouns and phrasal NPs, and that
integrating centering and informational constraints covers a broader
range of cases.  In \S\ref{data}, I present an analysis of all
discourse anaphoric NPs (N=1,233) in a corpus of ten narratives
showing that semantic explicitness depends largely on informational
constraints.  Discourse anaphoric NPs almost never provide new
information, and are rarely more informative than necessary.  In
\S\ref{model}, I show how Dale~\& Reiter's~\shortcite{dale&reiter94}
generation model can be augmented to apply uniformly to pronouns and
phrasal NPs for both generation and understanding.  While centering
has been used to account for informationally under-specified pronouns,
I argue that centering also accounts for certain over-specified
phrasal NPs.  In~\S\ref{integration}, I integrate centering with the
augmented Gricean model and discuss the extended coverage.  Results in
\S\ref{data} include a one-way correlation of overly informative
discourse anaphoric NPs with shifts in global discourse structure.  In
the conclusion, I discuss directions for extending the integrated
model in ways that might indirectly account for this correlation.

\section{Analysis of a Coded Corpus}
\label{data}

In this section, I present the results of an analysis of all discourse
anaphoric NPs in a corpus of spoken narratives directed at the
question of how informative NPs are, relative to their contexts of
occurrence.  The first subsection describes the corpus and coding
features. The next subsection presents results showing that discourse
anaphoric NPs in the corpus, whether pronominal or phrasal, are rarely
more informative than necessary, and if so, tend to occur at shifts in
global discourse structure.

Fig.~\ref{eg1} identifies four possibilities regarding the semantic
informativeness of an NP relative to its context.  Three of them
pertain to the following Gricean principles, referred to by
Dale~\shortcite{dale89} as informational adequacy and efficiency: the
speaker should be sufficiently informative to unambiguously identify
the intended referent (adequacy), and the speaker should be no more
informative than necessary (efficiency).  The boxed pronouns in (2a)
of Fig.~\ref{eg1} are both adequate and efficient (well-specified): it
is clear what the pronouns refer to; less informative forms (zero
pronouns) would be ungrammatical.  The phrasal NPs in (2b) are
adequate but not efficient (over-specified).  The pronominal NP in
(2c) is inadequate (under-specified; efficiency does not apply to
inadequate NPs): ``{\it it}'' could refer either to the ladder or the
tree. A fourth possibility is that an NP may perform two functions, to
identify the referent and to add information about it, as in (2d)
(over-determined). In Fig.~\ref{eg1}, the feature +/- increasing
distinguishes between over-determined and over-specified NPs.

{\scriptsize
\begin{figure}[t]
\begin{tabbing}
aaaa \= aaaaaaaaaaaaaaaaaaaaaaaaaaaaaaaaaaaaaaaa
\=aaaaaaaaaaaaaaaaaaaaaaaaaaa \kill
1 \> A man$_{1}$ saw a ladder$_{2}$ leaning against a pear tree$_{3}$. \\
2a. \> Later, \fbox{he$_{1}$} moved \fbox{it$_{2}$} to a different tree. \\
    \> $+$adequate; $+$efficient \> {\tt well-specified}\\
2b. \> \fbox{The man$_{1}$} moved \fbox{the ladder$_{3}$} to
                     a different tree. \\
    \> $+$adequate; $-$efficient; $-$increasing  \> {\tt over-specified}\\
2c. \> \fbox{It (?)} was tall.  \\
    \> $-$adequate \> {\tt under-specified}\\
2d. \> \fbox{The contented pear picker$_{3}$} was done for the day. \\
    \> $+$adequate; $+$increasing \> {\tt over-determined}
\end{tabbing}
\vspace{-.18in}
\caption{\label{eg1} Relative Informativeness}
\end{figure}
}

\subsection{Data Coding}

The corpus consists of ten narrations from Chafe's Pear
stories~\shortcite{chafe80}.  Chafe recorded and transcribed subjects who
had been asked to view the same movie and describe it to a second
person.  The movie contained seven sequential episodes about a man
picking pears. It had a vivid sound track, but no language.  As part
of a long term study of the relationship between linguistic features
and discourse structure~\cite{passonneau&litman93}
{}~\cite{litman&passonneau95a}~\cite{litman&passonneau95b},
discourse anaphoric NPs in the corpus had already been coded for
coreference relations and location.  Location of an NP is represented
here in terms of the containing sentential utterance and discourse
segment, as described below.  Fig.~\ref{excerpt} illustrates an excerpt.

{\scriptsize
\begin{figure}[h]
\begin{center}
\begin{tabular}{rrl}
\multicolumn{1}{c}{S$_{i}$} &
\multicolumn{1}{c}{U$_{j}$} \\
6  & 28   &  And you think "Wow, this little boy's$_{i}$ \\
   &      &  probably going to come and see the pears, \\
   & 29   &  [ps] and [ps] he$_{i}$ 's going to take
             a pear or two, \\
   &      &  and then.. go on his way."
\hspace{.5in} \\\hline
7  & 30   &  [ps] U-m but \fbox{the little boy$_{i}$} comes, \\
   & 31   &  [ps] a-nd u-h [1.0]] he$_{i}$  doesn't want
             just a pear, \\
   & 32   &  he$_{i}$  wants a whole basket. \\\hline
8  & 33   &  [ps] So \fbox{he$_{i}$} puts the- [ps] bicycle down,\\
   & 34   &  and he$_{i}$ [ps] you wonder \\
   &      & how he's$_{i}$ going to take it with this. \\
\end{tabular}
\caption{\label{excerpt} Narrative Excerpt Illustrating Informativeness}
\end{center}
\vspace{-.15in}
\end{figure}
}

Chafe~\shortcite{chafe80} identified three types of prosodic phrases from
graphic displays of intonation contours.  A period indicates a phrase
terminated by a pitch fall, a question mark indicates final level or
rising pitch, and a comma indicates phrase final---not
sentence-final---intonation.  The transcriptions here show all repeated
and incomplete words and phrases, non-lexical articulations such as
``uh, um, tsk'', and vowel lengthening as indicated by `-'.  Pause
locations are shown as `[ps]'.

Sentential utterances are defined be a non-overlapping sequence of
units that completely covers the discourse.  Briefly, a new sentential
utterance begins with a functionally independent clause (FIC) if it is
immediately adjacent to the preceding FIC.  Otherwise it begins at the
onset of the prosodic phrase where the next FIC begins.  An FIC is a
tensed clause that is not a verb argument, a restrictive relative
clause, or one of a set of formulaic ``interjection'' clauses (e.g.,
``{\it You know}'' with no clausal argument; for full details
cf.~\cite{passonneau94cod}).  Material between clauses includes
sentence or word fragments, and non-lexical articulations (e.g.,
``{\it um}'').  Locations and sequence numbers of the seven sentential
utterances in Fig.~\ref{excerpt} are shown in column 2.

The global context is structured into sequential segments,
multi-utterance units whose utterances are presumed to be more related
to one another semantically and pragmatically than to other
utterances.  The segments numbered 6-8 (col. 1 of Fig.~\ref{excerpt})
were derived from an empirical study described
in~\cite{passonneau&litman93}. Each narrative was segmented by 7 new,
untrained subjects.  Subjects were instructed to place segment
boundaries in transcripts whenever the narrator had finished one
communicative task and begun a new one.  They were restricted to
placing boundaries between prosodic phrases.  To focus their attention
on the criterion, subjects' were also instructed to label segments
with a brief description of the speaker's intention.

The size and number of segments per subject per narrative varied widely,
from a rate of 5.5\% to 41.3\% (Avg.=16\%), with segment widths ranging
from 1 to 49 phrases (Avg.=5.9).  Despite this variation, the number of
times 4 to 7 subjects assigned boundaries in the same place was
extremely significant (using Cochran's Q~\shortcite{cochran50};
cf.~\cite{passonneau&litman93}).  We took agreement among at least 4
subjects as the threshold for empirically validated boundaries.

A surface constituent is considered to be a discourse anaphoric NP if
it occurs in free variation with syntactically prototypical NPs, and
corefers with a preceding NP (cf.~\cite{passonneau94cod}).  One type
of empty category is also included, namely zero pronoun subjects of
FICs conjoined by ``,'', ``and'', etc.  In Fig.~\ref{excerpt}, the
sequence of coreferential NPs used to refer to the little boy are
coindexed.  Segments 7 and 8 in Fig.~\ref{excerpt} both begin with an
utterance containing an NP referring to the boy.  At the onset of
segment 7, a phrasal NP is used to refer to him (U$_{30}$) whereas at
the onset of segment 8 (U$_{33}$), a definite pronoun is used.  But
a pronoun could have replaced the phrasal NP in U$_{30}$ with no
loss of information.  So the phrasal NP is over-specified but
not over-determined; the attributes ``{\it boy}'' and ``{\it little}''
were already mentioned in U$_{28}$.  The pronoun subject in U$_{33}$
is locally well-specified because the boy is the only animate entity
mentioned in U$_{32}$; it is globally well-specified because the boy
is the only entity in the discourse with a bicycle.

\subsection{Analysis of Informational Constraints}
\label{analysis}

The goal of the analysis is to determine whether relative
informativeness of NPs correlates with global discourse structure
(cf.~\cite{reichman85}~\cite{gs86}).
Any phrasal NP that is discourse anaphoric is potentially
over-specified, whereas a definite pronoun will only be over-specified
if a zero pronoun could have been used.  I first sorted the discourse
anaphoric NPs in the corpus (N=1,233) into the three categories of
phrasal NPs (PhrNPs; N=563), explicit pronouns (PROs: definite,
indefinite, demonstrative; N=544), and zero pronominals (ZPs; N=126).
Then I identified all pairs of coindexed NPs where NP$_{2}$ was more
explicit than NP$_{1}$.  This procedure identified 128 discourse
anaphoric NPs in the corpus that were potentially over-specified or
over-determined.  The sole over-determined NP, illustrated in
Fig.~\ref{fig2}, occurs relatively late in the narrative (U$_{85}$);
it seems mainly to provide contrast (cf. ``{\it
that old man}'' vs. {\it those little boys}'').

{\scriptsize
\begin{figure}[t]
\begin{tabular}{rl}
\multicolumn{1}{c}{U$_{j}$} & \\
84 & [ps] You just know that those little boys are going to go back, \\
   & [ps] to where the pear tree was, \\
85 & and you just know \fbox{that old man}'s going to see [ps] these little\\
   & boys coming and say "Ha.. you're the ones who stole the pears."
\end{tabular}
\caption{\label{fig2} Over-determined NP}
\end{figure}
}

Potentially over-specified NPs were sorted into four mutually exclusive
categories---well-specified, segment onset, attentional shift, and
reiterative.  A potentially over-specified NP is well-specified if a
less explicit form would have been ambiguous or unclear.  The containing
utterance is included in the context since the proposition expressed in
an utterance can disambiguate a referring expression.  A potentially
over-specified NP that is not well-specified, but which occurs in the
first utterance of a new segment, is classified as a segment onset.  The
segments in the coded Pear corpus arguably contain intra-segmental
shifts of attention associated with changes in temporal aspect, or
shifts in discourse reference time (for definitions assumed here,
cf.~\cite{kameyama&etal93}). The third category, attentional shift,
consists of these cases.  A fourth catch-all category includes, e.g.,
repetitions, repairs, contrastive NPs and unexplained cases.

{\scriptsize
\begin{table}[b]
\begin{center}
\begin{tabular}{|lrrrrr|}\hline\hline
\multicolumn{1}{|c}{Antecedent} & \multicolumn{1}{c}{Well-} &
\multicolumn{1}{c}{Segment} &
\multicolumn{1}{c}{Atten.} &
\multicolumn{1}{c}{Other} &
\multicolumn{1}{c|}{Total} \\
\multicolumn{1}{|c}{Segment} & \multicolumn{1}{c}{Specified} &
\multicolumn{1}{c}{Onset} &
\multicolumn{1}{c}{Shift} & & \\\hline
Same     &    22   &    -   &   21     &   15   &   58 \\
\%                  &    38\% &    -   &   36\%   &   26\% &  100\% \\\hline
Prev     &    37   &   20   &   10     &    4   &   69 \\
\%                  &    53\% &   29\% &   12\%   &    6\% &  100\% \\\hline
Totals              &    59   &   20   &   29     &   19   &  127 \\
\%                  &    46\% &   16\% &   23\%   &   15\% &  100\% \\\hline
\end{tabular}
\end{center}
\caption{\label{table1}  Potentially Over-Specified NPs}
\end{table}
}

Table~\ref{table1} indicates that most potentially over-specified NPs
(N=127) were either well-specified (46\%) or occur at an empirically
verified segment onset (16\%) or a hypothesized attentional shift
(23\%).  Of the 69 NPs whose nearest antecedent was in a distinct
segment, 29\% occurred at a segment onset.  Over a third (36\%) of the
NPs whose antecedent was in the same segment, and 12\% of those whose
antecedent was in a distinct segment occurred at an intra-segmental
attentional shift. In sum, in the coded Pear corpus, NPs that re-evoke
existing entities seem to be rarely over-specified (68/1233, or 5.5\%)
or over-determined (1/1233).  Of the 68 over-specified cases (columns
2-5), 20 (30\%) correlate with segment onsets independently identified
by naive subjects, and 29 (42\%) appear to correlate with
intra-segmental attentional shifts.  Thus, an over-specified NP is
more likely than not to correlate with an attentional shift (72\%).
Note however, that the reverse implication does not hold, that is, it
is not the case that a segment shift is likely to be signalled by an
over-specified NP.

\subsection{Focused Attribute Sets}
\label{fav}

To account for the choice of modifiers in phrasal discourse anaphoric
NPs, it is necessary to determine how attributes are selected from the
information known about a discourse entity.  According to
Grice's~\shortcite{grice75} maxim of quality, speakers should be relevant.
With respect to discourse anaphoric NPs in the Pear stories, NP
modifiers are derived from what I refer to as focussed attribute sets,
independent of whether the NP is over-specified.  Focussed attribute
sets comprise the following three categories of relevance.  First, an
attribute set can be in focus because it was mentioned in the most
recent phrasal NP.  For example, in Fig.~\ref{excerpt}, the boy is
referred to in U$_{30}$ as ``{\it the little boy},'' repeating
attributes mentioned in the last phrasal NP referring to the boy (in
U$_{28}$).

Second, the focussed attribute set may specify the most recently
mentioned location of an entity.  The subject NP in U$_{17}$ of
Fig.~\ref{fig2prime} (\S\ref{c-describe}) refers to one man as ``{\it
the man up in the tree}'' to distinguish him from the second man who
came by with a goat.  The tree is the last mutually known location of
the former.  Finally, an attribute set can be in focus because it
pertains to a key narrative event that the entity has been an agent
of.  Intuitively, an event is more central to a narrative the more
difficult it is to describe the narrative without mentioning that
event.  Operationally, key events occur more frequently than others
both within and across narratives.  For example, the main adult
character is often described as ``{\it the pear picker},'' or as
``{\it the man who was picking pears}'' (see U$_{108}$ of
Fig.~\ref{fig3}, \S\ref{integration}), and so on; the other main
character is often described as ``{\it the thief},'' ``{\it the boy
who stole the pears},'' ``{\it the boy with the pears},'' and so on.

How to order the focussed attribute sets for a given discourse entity
is a topic for further investigation. Here, I simply assume that the
three types of attribute sets mentioned above---where applicable---are
in focus.  I also assume that the focussed attribute sets of an entity
(FAV$_{e}$) are updated as the discourse progresses.

\section{Modelling Informativeness of NPs}
\label{model}

The data reported above indicates that in the Pear corpus, definite
pronouns and phrasal NPs are rarely over-specified or over-determined.
In this section, I describe a processing model to account for this
observation.  In the next section, I discuss how centering can be
integrated with this model to account for under-specified pronouns,
and certain over-specified phrasal NPs.  First, I briefly review
Dale's~\shortcite{dale89}~\shortcite{dale92} model, including his more recent
work with Reiter~\shortcite{dale&reiter94}.  Then I modify this model to
apply to understanding as well as generation; to include the current
utterance in the context of evaluation; to apply informational
constraints uniformly to pronouns and phrasal NPs; and to select
modifiers on the basis of focused attribute-value pairs.

\subsection{Distinguishing Descriptions}
\label{dd}

Dale~\shortcite{dale89} generates anaphoric
pronouns and phrasal NPs by distinct means.  In
EPICURE~\shortcite{dale89}, a system for generating
recipes, a definite pronoun is always generated to refer to the
`discourse center', which is analogous to the backward-looking center
of~\cite{gjw83}~\cite{kameyama85}, but is domain specific. It is the
entity that results from the next recipe operation.  For example,
rice$_{1}$ will be the center after an utterance of {\it Stir the
rice$_{1}$}.

Dale~\shortcite{dale89} requires phrasal NPs to be
distinguishing descriptions.  As in Webber~\shortcite{webber78}, Dale
assumes that the discourse model represents the discourse entities
that have already been evoked, and the attribute-value pairs
describing them.  For any set of entities U,
Dale~\shortcite{dale89} defines a distinguishing
description of an entity $e$ in U to be a set of attribute-value pairs
that are true of $e$, and of no other members of U.  This enforces
adequacy.  He defines a minimal distinguishing description to be one
where the cardinality of the attribute-value pairs cannot be reduced.
This addresses efficiency.\footnote{Cf.  Reiter~\shortcite{reiter90}
for a discussion of problems in generating maximally efficient NPs
using Dale's framework, and Dale~\& Reiter~\shortcite{dale&reiter94}
for an argument that maximal efficiency is psychologically
implausible.}

Dale~\shortcite{dale89} defines the discriminatory
power (${\cal F}$) of an attribute-value pair $<$A, V$>$ that is true
of a discourse entity $e$ in a universe of entities U in terms of the
cardinality $N$ of U, and the total number $n$ of entities in U that
$<$A, V$>$ is true of:

${\cal F}(<A, \: V>, \: U) \: = \: \frac{N \: - \: n}{N \: - \: 1}$

\noindent
${\cal F}$ ranges in value from 0 to 1.  If $<A, \: V>$ is true of only
one of the entities in the set U, then ${{\cal F}_{<A, \: V>}}$ is 1,
and $<A, \: V>$ is a distinguishing description of the entity.

Dale's~\shortcite{dale89} algorithm for
constructing a distinguishing description of $e$ in U, given a set
${\cal P}$ of attribute-value pairs that are true of $e$, briefly
works as follows.  First compute ${\cal F}$ for each member of ${\cal
P}$.  If all values of ${\cal F}$ are 0, no unique description can be
constructed.  Otherwise, select the attribute-value pair with the
highest value to add to the description, and reset U to be only those
entities in the initial U that the selected attribute-value pair is
true of.  Repeat this process, terminating when an attribute-value
pair with a discriminatory power of 1 has been selected.  The selected
attribute-value pairs constitute the input description for a surface
NP.

In recent work, Dale~\& Reiter~\shortcite{dale&reiter94} enforce a
range of Gricean constraints using an algorithm based on human
behavior that is simpler and faster than their previous
algorithms~\cite{dale89}~\cite{reiter90}. It performs
less length-oriented optimization, thus balancing brevity against
lexical preference.  The output NPs are not guaranteed to be maximally
short because humans occasionally use unnecessary modifiers.  The
5.5\% rate of over-specified discourse anaphoric NPs in the Pear data
also supports the relaxation of brevity, but is partly conditioned by
attentional factors (cf.\S\S\ref{integration}-\ref{conclusion}).

\subsection{C\_describe}
\label{c-describe}

In this section I illustrate the role of $c\_describe$ in processing
definite pronouns and phrasal NPs.
C\_describe is a 4-place relation among a discourse entity E, a
surface NP, the current utterance context $\lambda U$, and the
discourse context C that requires $\lambda NP \lambda U$ to be a
distinguishing description of E relative to C.  For generation, NP is
solved for given an instantiation of the remaining three arguments,
whereas E is solved for during understanding (assuming Prolog's
control structure).

A definite pronoun that is a distinguishing description is also a
minimal distinguishing description because its length is 1. In
generation, C$\_describe$ attempts first to find a definite pronoun to
satisfy the uninstantiated NP argument, succeeding if the pronoun is a
distinguishing description.  For generating the pronoun ``{\it he}''
in U$_{4}$ of Fig.~\ref{fig6}, the arguments of $c\_describe$ are:

\begin{eqnarray}
\label{u-4}
\lefteqn{c\_describe(e_{1}, \: NP,}\\
\nonumber \lefteqn{\hspace{.15in}
\lambda X \: fills'(X,``his \: thing'',``pears''), \: FS_{1})}
\end{eqnarray}

\noindent
The utterance context is assumed to be a feature structure co-indexed
with any relevant discourse entities other than the uninstantiated
variable E.\footnote{For simplicity, the utterance context represents
certain semantic arguments as quoted strings.}  By using the utterance
as part of the input in solving for NP, given information that appears
anywhere in the current utterance can filter entities from the
discourse context, following
Dale's~\shortcite{dale89} algorithm.  New
information about an entity in the utterance is not mutually known,
and has no discriminatory power~\cite{dale89}.

For present purposes, the last argument of $c\_describe$ is first
instantiated to the most recent focus space, and in turn to other
focus spaces until a solution is found.
Dale~\shortcite{dale89} takes the universe of
discourse to be partitioned into focus spaces (cf.~\cite{gs86}), with
the most recent focus space being the most accessible, and making no
assumptions regarding relative accessibility of earlier focus spaces.
Similar assumptions are made here.  I assume that segment boundaries
in the Pear corpus correspond to focus spaces, and that some focus
spaces may be composed of others. I assume the existence of an
inference mechanism that constrains how focus spaces are signalled
during generation, and how focus spaces are inferred during
understanding.  In recent work, for example, Litman and I report on
algorithmic methods for identifying segment boundaries in the Pear
corpus using features of prosody, cue words and referential
NPs~\cite{litman&passonneau95b}.  Given such a mechanism, a new focus
space would be added to the discourse model after a
segment onset has been processed.

{\scriptsize
\begin{figure}[t]
\begin{tabular}{rrl}
\multicolumn{1}{c}{S$_{i}$} &
\multicolumn{1}{c}{U$_{j}$} \\
2 &  4 & [ps] A- nd \fbox{he (e$_{1}$)} [ps] fills his- thing with pears, \\
  &  5 & and ZERO (e$_{1}$) comes down \\
  &  6 & and there's a basket he (e$_{1}$) puts them in.  \\\hline
3 &  7 & [ps] A-nd you see- [ps] passerbyers on  \\
  &    &      \hspace{.18in} bicycles and stuff go by.  \\
  &  8 & [ps] A-nd [ps] then a boy (e$_{2}$) comes by, \\
  &    & [ps] on a bicycle, \\
  &  9 & \fbox{the man (e$_{1}$)} is in the tree,\\
  & 10 & [ps] and \fbox{the boy (e$_{2}$)} gets off the bicycle, \\
  & 11 & and ZERO (e$_{2}$) looks at the man (e$_{1}$) , \\
\end{tabular}
\caption{\label{fig6} Excerpt from Narrative 9}
\end{figure}
}

In (\ref{u-4}), FS$_{1}$ appears as the initial context argument of
$c\_describe$.  The only animate entity in FS$_{1}$ is e$_{1}$,
previously described as a man picking pears in a pear tree who looks
like a farmer, is plump, has a mustache, and is wearing a white apron
(utterances 1-3, not shown here).  The feature structures
corresponding to all but one of the definite pronouns ``{\it he, she,
it}'' or ``{\it they}'' will be rejected as a description of e$_{1}$
because e$_{1}$ is neither plural, non-animate or female.\footnote{For
simplicity, I am ignoring the difference between grammatical gender
and sex.} The pronoun ``{\it he}'', represented as the attribute-value
pairs ($<$type: human$>$, $<$gender: male$>$, $<$cardinality: 1$>$),
not only describes e$_{1}$, it is also a minimal distinguishing
description.

An analogous process applies to understanding the same pronoun in
U$_{4}$, with the entity variable E uninstantiated, NP instantiated to
``{\it he}'', and the utterance and discourse context instantiated as
above.  Given a distinguishing description, there is
guaranteed to be exactly one solution to E. However, the search
problem increases with the size of the context.  Partitioning the
search space into focus spaces controls the search through the
discourse model to some degree.  (Integrating centering with
$c\_describe$ as described below guides the search even
further.)  For present purposes, $c\_describe$ returns E instantiated
to e$_{1}$ after searching through the entities in FS$_{1}$.
The remaining NPs exemplified here are understood in a similar fashion.

Given a context where there is no definite pronoun solution to NP,
$c\_describe$ will attempt to construct a phrasal NP, preferably with
no modifiers.  In Fig.~\ref{fig6}, a new, singular, male, human entity
is added to the context at U$_{8}$: a boy who comes by on a bicycle
(e$_{2}$).  Subsequent references to the boy or the man must
discriminate between them.  The utterance context for the subject NP
of U$_{9}$---$\lambda \: X \: in'(X,``the
\; tree'')$---does not identify e$_{1}$ because U$_{5}$---``{\it he
comes down}''---leads to the inference that the man is no longer in
the tree.  However, e$_{1}$ is a male adult and e$_{1}$ is a male
child, a distinction encoded by the common nouns ``{\it man}'' versus
``{\it boy}''.  Since ``{\it man}'' is what Dale~\&
Reiter~\shortcite{dale&reiter94} refer to as a basic attribute, ``{\it
man}'' will be selected as the head noun.  The determiner will be
definite because the entity is already in the context (but
cf.~\cite{passonneau94foc}).  The resulting NP {\it the man} is a
minimal distinguishing description because no pronoun is
a distinguishing description.

{\scriptsize
\begin{figure}[t]
\begin{tabular}{rrl}
\multicolumn{1}{c}{S$_{i}$} &
\multicolumn{1}{c}{U$_{j}$}  \\
  & 13 & and it's just a monotonous kind of thing for him (e$_{1}$). \\\hline
4  & 14 & [ps] And a man (e$_{2}$) comes along with a goat,   \\
   & 15 & [ps] and the goat obviously is interested in the pears.  \\
   & 16 & But the man (e$_{2}$) just walks by with the goat.   \\\hline
5  & 17 & And \fbox{the man (e$_{1}$) up in the tree}\\
   &      & \hspace{.12in} doesn't even notice.
\end{tabular}
\caption{\label{fig2prime}  Phrasal NP to Avoid Ambiguity}
\end{figure}
}

Fig.~\ref{fig2prime} illustrates a context where a phrasal NP without
modifiers could not both have a head noun that specifies a basic
attribute, and be a distinguishing description.  It also illustrates
the problematic nature of relations among distinct focus spaces.  In
generating the subject NP in U$_{17}$, the last argument of
$c\_describe$ is first instantiated to FS$_{4}$.  The pears referred
to in U$_{15}$ of segment 4 are some pears that e$_{1}$ picked, so in
order to interpret U$_{15}$, e$_{1}$ must be brought into focus.  This
side-effect of resolving the reference to the pears could be
implemented by adding e$_{1}$ to FS$_{4}$, or by resetting the current
focus space to a more encompassing focus structure that includes
FS$_{3}$ and FS$_{4}$.  I believe further empirical work is needed to
resolve such issues.  In any case, I assume that the context for
generating U$_{17}$ includes both e$_{1}$ and e$_{2}$.  Because these
two entities are the same type, a distinguishing description of
e$_{1}$ must contain discriminatory modifiers.  Features for
generating the modifiers are selected from FAV$_{e_{1}}$, which here
contains only two sets of salient attributes. Since e$_{1}$'s location
is the most recently evoked, it is used in generating the NP.

Above, I noted that centering can add structure to the search space
for understanding discourse anaphoric NPs.  Fig.~\ref{fig6}
illustrates another reason to integrate centering with $c\_describe$.
In U$_{10}$ of Fig.~\ref{fig6}, the subject NP (``{\it the boy}'') is
not a pronoun even though the utterance context is a distinguishing
description of e$_{2}$.  The boy (e$_{2}$) is mutually known to have
been on a bicycle at the time of the event mentioned in utterance
U$_{8}$.  Temporal processing (cf.~\cite{kameyama&etal93}) would lead
to the inference that the boy is still on the bicycle after U$_{9}$.
Thus a definite pronoun is presumably well-specified, and the model
presented so far would generate ``{\it he}''.  However, a pronoun
would produce a garden path effect in this context; i.e., it would be
interpreted as referring to the man until ``{\it bicycle}'' has been
processed.
\section{Centering and Informativeness}
\label{integration}

The $c\_describe$ relation has three limitations that centering can
compensate for.  First, $c\_describe$ constrains the semantic content
of a discourse anaphoric NP, but not its grammatical role.  Second, as
noted below, centering predicts that a pronoun can be under-specified.
Third, an explanation is needed for the over-specified NP {\it the
boy} in U$_{10}$ of Fig.~\ref{excerpt}.  In this section, I indicate
how centering is interleaved with $c\_describe$.  Centering is a more
local process so it applies first.

\subsection{Centering}
\label{centering}

Centering is a model of local focus of attention that constrains the
use of definite pronouns~\cite{gjw83}~\cite{kameyama85}.  One of the
discourse entities~\cite{webber78} evoked by an NP in an utterance
U$_{i}$ may be the backward-looking center (CB)~\cite{gjw83} of
U$_{i}$, the current local focus of attention.  Alternatively, the CB
of U$_{i}$ (CB$_{U_{i}}$) might not be explicitly mentioned (realized)
in the utterance~\cite{gjw83}.  The discourse entities mentioned in
U$_{i}$ comprise the forward looking centers (CFs), ordered by
increasing obliqueness of grammatical
role~\cite{kameyama85}~\cite{passonneau89} to represent the likelihood
that they will be mentioned in the subsequent utterance.  The
centering principle~\cite{gjw83} predicts that if CB$_{U_{i}}$ and
CB$_{U_{i-1}}$ are the same entity, then the NP evoking CB$_{U_{i}}$
will be a third person, definite pronoun.

\addtocounter{example}{1}
\bex
\label{egg}
a. Carmella$_{j}$ went to the bookstore. \\
b. Afterwards, she$_{j}$ gave Rachel$_{k}$ a new book. \\
c. She$_{j}$'s a true bibliophile.
\eex

Example (\ref{egg}) illustrates that where the semantics of the
utterance and commonsense reasoning do not discriminate among possible
referents for an ambiguous pronoun, there is an independent effect of
local attentional constraints.  Centering predicts that the preferred
interpretation of the pronoun in (\ref{egg}c) is Carmella.  But in this
context, neither the pronoun alone nor the utterance is a distinguishing
description of anyone, so the pronoun is under-specified.

\bex
\label{lmn}
a. Carmella$_{j}$ went to the bookstore. \\
b. On her way home, she$_{j}$ saw Rachel$_{k}$. \\
c. She$_{k}$ looked pale.
\eex

Kameyama~\shortcite{kameyama85} used examples like (\ref{lmn}) to
illustrate how commonsense reasoning and lexical semantics can
override the default centering predictions for pronoun interpretation.
Centering would predict that `Carmella' is the backward-looking center
of (\ref{lmn}b), and that the default interpretation of the pronoun in
(\ref{lmn}c) would thus be `Carmella'.  Instead, (\ref{lmn}c) is
interpreted as a continuation of the description of the perceptual
event in (\ref{lmn}b).  Kameyama~\shortcite{kameyama86} posits property
sharing of features of adjacent utterances as a constraint on CB,
where the shared property can be subject (or non-subject) grammatical
role (cf.~\cite{passonneau89}), as in (\ref{egg}), or what she refers
to as empathy, as in (\ref{lmn}).  Note that because `Rachel' is
already known to be the object of the perceptual event in
(\ref{lmn}b), the utterance context in (\ref{lmn}c) is a
distinguishing description of `Rachel.'

\subsection{Integrated Model}
\label{final-model}

{\scriptsize
\begin{figure}[b]
\begin{tabular}{rrl}
\multicolumn{1}{c}{S$_{i}$} &
\multicolumn{1}{c}{U$_{j}$} & \\
21 & 105 & [ps] So they (e$_{1}$)'re walking along, \\
   & 106 & and they (e$_{1}$) brush off
          their pears (e$_{3}$),  \\
   & 107 & and they (e$_{1}$) start
         eating it (e$_{3}$). \\\hline
22 & 108 & Then \fbox{they (e$_{1}$)} walk by-[ps] \\
   &     &  \fbox{the man who was picking the pears (e$_{2}$)} \\
\end{tabular}
\caption{\label{fig3} Excerpt from Narrative 1}
\end{figure}
}

Fig.~\ref{fig3} shows all of one segment and part of another one where
the subject pronouns of all the utterances are coreferential.  On the
one hand, the CB of the segment initial utterance U$_{108}$ is the
same as the CB of U$_{107}$, conflicting with the idea expressed
in~\cite{gjw86} that centering transitions reflect global discourse
coherence (cf.~\cite{passonneau94a}).  On the other hand, integrating
centering and $c\_describe$ can account for both NPs in U$_{108}$
and support inferences consistent with a global focus shift.

Earlier in the narrative excerpted in Fig.~\ref{fig3}, three boys helped
the pear thief after he had fallen off of his bicycle, and were
rewarded with three pears.  Segment 21 describes their adventures
after the pear thief leaves.  In generating utterance U$_{108}$, the
input to the generator will be a representation of an event in which
the boys eat their pears. The set of three boys is designated as the
new CB.  Because CB$_{U_{108}}$ is the same as CB$_{U_{107}}$, it
should be realized as a pronoun~\cite{gjw83}, and by property
sharing~\cite{kameyama85}~\cite{passonneau89}, it should be realized
as the subject of the current utterance.  Centering and Gricean
constraints coincide here in that the definite pronoun ``{\it they}''
is also a minimal distinguishing description.

To generate the phrasal NP object in U$_{108}$, the process is
analogous to that discussed above for generating ``{\it the man up in
the tree}'' in Fig.~\ref{fig2prime}. The context argument of
$c\_describe$ is first set to Cf$_{U_{107}}$.  Since neither
Cf$_{U_{107}}$ nor the most accessible focus space (FS$_{21}$)
contains a representation of e$_{2}$, the context argument will be
reset until e$_{2}$ is in a focus space on the focus stack.  Focussed
attribute sets are then used to generate the relative clause.

For understanding the subject NP in U$_{108}$, recall that centering
applies before $c\_describe$.  The subject pronoun will be assumed to
realize the CB of the utterance, and will be assigned the default
interpretation of e$_{1}$.  Application of $c\_describe$ leads to the
recognition that ``{\it they}'' is also a distinguishing description
of e$_{1}$ relative to CF$_{U_{107}}$. In this fashion, centering
prunes the search space to the single entity satisfying the
informational constraints imposed by $c\_describe$.  In understanding
the object NP, the context argument must be instantiated to a more
inclusive focus space, since neither the previous utterance nor the
previous segment contains any entities described by this NP.

The integrated model also accounts for the problematic phrasal NP
in Fig.~\ref{fig6}, discussed above.  We saw that for U$_{9}$ and
U$_{10}$, repeated below, the phrasal subject of U$_{9}$ was
well-specified, but the phrasal subject of U$_{10}$ was
over-specified, and a pronoun would be generated instead. But
as noted above, a pronoun subject would have a garden path effect.

\begin{description}
\item[U$_{9}$: \hspace{.005in}]  the man (e$_{1}$) is in the tree (e$_{3}$),
\vspace{-.02in}
\item[U$_{10}$:] and the boy (e$_{2}$) gets off the bicycle (e$_{4}$),
\end{description}

\noindent
Kameyama's version of centering~\shortcite{kameyama86} differs
from~\cite{gjw83} in allowing an utterance to have a null CB.
U$_{10}$ would have a null CB because there is no definite pronoun
constrained by property sharing that corefers with an NP in the
previous utterance; in fact no NPs in U$_{10}$ refer to entities
mentioned in U$_{9}$.  A definite pronoun subject in U$_{10}$ would be
assumed to be CB$_{U_{10}}$ and would be inferred to refer to e$_{1}$.
This accounts for the garden path effect.  Consequently, a pronoun
must be blocked.  Because no entity in U$_{9}$ is referred to in
U$_{10}$, the input for generating U$_{10}$ will be annotated as
having a NULL CB. This imposes output constraints requiring the
subject and object NPs to be other than definite pronouns.  As a
consequence, $c\_describe$ will not try to find a pronoun solution to
the uninstantiated NP argument.  In the first phrasal NP solution, the
head would denote a basic category and the NP would have no
modifiers, thus generating the existing phrase ``{\it the boy}''.  In
sum, centering relaxes the constraint otherwise imposed by
$c\_describe$ that an NP cannot be over-specified.

\section{Conclusion}
\label{conclusion}

I have presented an analysis of discourse anaphoric phrasal NPs in a
corpus of narrative monologues showing that pronouns and phrasal NPs
are rarely over-specified.  Future research should indicate to what
degree this generalization applies to other genres and modalities.
Centering predicts conditions under which an under-specified pronoun
can be used, but says little about the interpretation of phrasal NPs.
I have outlined a processing model that integrates the attentional
constraints of centering with aspects of Grice's maxims of quantity
and quality.  For enforcing the maxim of quantity, I rely on Dale's
algorithm for constructing distinguishing
descriptions~\shortcite{dale89}~\shortcite{dale92},
which I apply uniformly to
pronouns and phrasal NPs for both generation and understanding.  For
enforcing the maxim of quality, I combine aspects of Dale~\&
Reiter's~\shortcite{dale&reiter94} preferred attributes with the construct
of focussed attribute sets derived from the corpus analysis.  In
contrast to Dale~\& Reiter~\shortcite{dale&reiter94}, distinguishing
descriptions are evaluated using the current utterance context as a
filter, and by instantiating the discourse context successively to the
Cf list of the preceding utterance, then the current focus space, then
other focus spaces, until a solution is found.

Centering provides one mechanism for relaxing the requirement that an
NP (either pronominal or phrasal) should be a distinguishing
description.  Another mechanism would be needed to relax informational
constraints at shifts in focus structure, so as to account for the
one-way implication of over-specified NPs with global shifts of
attention (Table~\ref{table1}).  However, further investigation is
needed to determine how to integrate local and global discourse
processing.  When neither the Cf list nor the current focus space is
the appropriate context for understanding or generating a discourse
anaphoric NP, I have assumed that either an earlier focus space or a
more inclusive one must be accessed.  Some of the examples presented
here suggest that the contextual dependencies captured by the use of
focused attributes might constrain the relation of each new utterance
to the global discourse model.  For example, the segment onset in
Fig.~\ref{fig3} (U$_{108}$) contains two NPs, one of which is the same
as the CB of the preceding utterance.  Maintaining the same CB relates
U$_{108}$ and its focus space (FS$_{22}$) to the most recent focus
space FS$_{21}$.  But the object NP expresses attributes last
mentioned in segment 17, thus relating U$_{108}$ to the earlier focus
space FS$_{17}$.  If the global structure is a tree, the relation of
U$_{107}$ to both segments 21 and 17 might indicate how high up in the
tree to locate the new focus space.  Alternatively, an investigation
of such relations might provide evidence about the nature of global
structure, such as whether it is a tree or a lattice.

\section*{Acknowledgements}
{\footnotesize
This work was partly supported by NSF grant IRI-91-13064.
Thanks to Robert Dale, Ehud Reiter, and Megumi Kameyama
for valuable comments on ideas presented here.}

{\small
\bibliographystyle{named}

\begin{thebibliography}{}

\bibitem[\protect\citeauthoryear{Chafe}{1980}]{chafe80}
W.~L. Chafe.
\newblock {\em The Pear Stories}.
\newblock Ablex Publishing Corporation, Norwood, NJ, 1980.

\bibitem[\protect\citeauthoryear{Cochran}{1950}]{cochran50}
W.~G. Cochran.
\newblock The comparison of percentages in matched samples.
\newblock {\em Biometrika}, 37:256--266, 1950.

\bibitem[\protect\citeauthoryear{Dale and Reiter}{To Appear}]{dale&reiter94}
R.~Dale and E.~Reiter.
\newblock Computational interpretations of the {G}ricean maxims
  in the generation of referring expressions.
\newblock To appear in Cognitive Science.

\bibitem[\protect\citeauthoryear{Dale}{1989}]{dale89}
R.~Dale.
\newblock Cooking up references.
\newblock In {\em Proceedings of the 27th Annual ACL}, 1989.

\bibitem[\protect\citeauthoryear{Dale}{1992}]{dale92}
R.~Dale.
\newblock {\em Generating Referring Expressions.}.
\newblock MIT Press, Cambridge, MA, 1992.

\bibitem[\protect\citeauthoryear{Grice}{1975}]{grice75}
H.~P. Grice.
\newblock Logic and conversation.
\newblock In {\em Syntax and Semantics, Vol. 3: Speech Acts}, pages 41--58.
  Academic Press, New York, 1975.

\bibitem[\protect\citeauthoryear{Grosz and Sidner}{1986}]{gs86}
B.~J. Grosz and C.~L. Sidner.
\newblock Attention, intentions and the structure of discourse.
\newblock {\em Computational Linguistics}, 12:175--204, 1986.

\bibitem[\protect\citeauthoryear{Grosz \bgroup \em et al.\egroup
  }{1983}]{gjw83}
B.~J. Grosz, A.~K. Joshi, and S.~Weinstein.
\newblock Providing a unified account of definite noun phrases in discourse.
\newblock In {\em Proceedings of the 21st ACL}, 1983.

\bibitem[\protect\citeauthoryear{Grosz \bgroup \em et al.\egroup
  }{1986}]{gjw86}
Barbara~J. Grosz, A.~K. Joshi, and S.~Weinstein.
\newblock Towards a computational theory of discourse interpretation, 1986.
\newblock Ms.

\bibitem[\protect\citeauthoryear{Kameyama \bgroup \em et al.\egroup
  }{1993}]{kameyama&etal93}
M.~Kameyama, R.~J. Passonneau, and M.~Poesio.
\newblock Temporal centering.
\newblock In {\em Proceedings of the 31st ACL}, 1993.

\bibitem[\protect\citeauthoryear{Kameyama}{1985}]{kameyama85}
M/ Kameyama.
\newblock {\em Zero Anaphora: The Case of Japanese}.
\newblock PhD thesis, Stanford University, 1985.

\bibitem[\protect\citeauthoryear{Kameyama}{1986}]{kameyama86}
M. Kameyama.
\newblock A property-sharing constraint in centering.
\newblock In {\em Proceedings of the 24st ACL}, 1986.

\bibitem[\protect\citeauthoryear{Litman and
  Passonneau}{1995}a]{litman&passonneau95b}
D.~Litman and R.~Passonneau.
\newblock Combining multiple knowledge sources for discourse segmentation.
\newblock In {\em Proceedings of the 33rd ACL}, 1995.

\bibitem[\protect\citeauthoryear{Litman and
  Passonneau}{1995}b]{litman&passonneau95a}
D.~Litman and R.~Passonneau.
\newblock Developing algorithms for discourse segmentation.
\newblock In {\em Working Notes of AAAI Spring Symposium Series on Empirical
  Methods in Discourse Interpretation and Generation}, 1995.

\bibitem[\protect\citeauthoryear{Passonneau and
  Litman}{1993}]{passonneau&litman93}
R.~J. Passonneau and D. Litman.
\newblock Intention-based segmentation: Reliability and correlation with
  linguistic cues.
\newblock In {\em Proceedings of the 31st ACL}, 1993.

\bibitem[\protect\citeauthoryear{Passonneau and
  Litman}{To appear}]{passonneau&litman94}
R.~J. Passonneau and D. Litman.
\newblock Empirical Analysis of Three Dimensions of Spoken
	Discourse: Segmentation, Coherence and Linguistic Devices.
\newblock In E. Hovy and D. Scott, eds., {\em Burning Issues in Discourse}.
Springer Verlag, To appear.

\bibitem[\protect\citeauthoryear{Passonneau}{1989}]{passonneau89}
Rebecca~J. Passonneau.
\newblock Getting at discourse referents.
\newblock In {\em Proceedings of the 27th ACL}, 1989.

\bibitem[\protect\citeauthoryear{Passonneau}{1994}a]{passonneau94cod}
R.~J. Passonneau.
\newblock Protocol for coding discourse referential noun phrases \\
and their antecedents.
\newblock Technical report, Columbia University, 1994.

\bibitem[\protect\citeauthoryear{Passonneau}{1994}b]{passonneau94foc}
R.~J. Passonneau.
\newblock Frame shifts: Perlocutionary meaning and discourse reference.
\newblock In {\em Conference on Focus and Natural Language Processing},
  Kassel,  Germany, 1994.

\bibitem[\protect\citeauthoryear{Passonneau}{To appear}]{passonneau94a}
R.~J. Passonneau.
\newblock Interaction of the segmental structure of discourse with
  explicitness of discourse anaphora.
\newblock In E. Prince, A. Joshi, and M. Walker, eds., {\em
  Proceedings of the Workshop on Centering Theory in Naturally Occurring
  Discourse}. Oxford University Press, To appear.

\bibitem[\protect\citeauthoryear{Reichman}{1985}]{reichman85}
R.~Reichman.
\newblock {\em Getting Computers to Talk Like You and Me}.
\newblock MIT Press, Cambridge, Massachusetts, 1985.

\bibitem[\protect\citeauthoryear{Reiter}{1990}]{reiter90}
E. Reiter.
\newblock {\em Generating appropriate natural language object descriptions}.
\newblock PhD thesis, Harvard University, 1990.

\bibitem[\protect\citeauthoryear{Sidner}{1979}]{sidner79}
C.~L. Sidner.
\newblock Towards a computational theory of definite anaphora
  comprehension in{E}nglish discourse.
\newblock Technical report, MIT AI Lab, 1979.

\bibitem[\protect\citeauthoryear{Webber}{1978}]{webber78}
B.~Webber.
\newblock A formal approach to discourse anaphora.
\newblock Technical Report 3761, Bolt Beranek and Newman Inc., 1978.

\end{thebibliography}

}
\end{document}